\pdfoutput=1
\documentclass[11pt]{article}

\usepackage[]{acl}

\usepackage{times}
\usepackage{stfloats}
\usepackage{latexsym}
\usepackage{multirow}
\usepackage{amsmath}
\usepackage{bbding}
\usepackage{booktabs}
\usepackage[accsupp]{axessibility} 

\usepackage[T1]{fontenc}
\usepackage[utf8]{inputenc}

\usepackage{microtype}

\usepackage{inconsolata}

\usepackage{graphicx}

\title{Hallucination Mitigating for Medical Report Generation}

\author{
{\large \bf Ruoqing Zhao, Runze Xia, Piji Li\thanks{Corresponding author.}} \\
\texttt{swieo25@gmail.com,xiarunze@nuaa.edu.cn, pjli@nuaa.edu.cn}\\
 College of Artificial Intelligence, Nanjing University of Aeronautics and Astronautics, Nanjing, China \\
 MIIT Key Laboratory of Pattern Analysis and Machine Intelligence, Nanjing, China \\
 The Key Laboratory of Brain-Machine Intelligence Technology, Ministry of Education, Nanjing, China \\
}

\begin{document}
\maketitle
\begin{abstract}
In the realm of medical report generation (MRG), the integration of natural language processing has emerged as a vital tool to alleviate the workload of radiologists. Despite the impressive capabilities demonstrated by large vision language models (LVLMs) in understanding natural language, their susceptibility to generating plausible yet inaccurate claims, known as ``hallucinations'', raises concerns-especially in the nuanced and critical field of medical. In this work, we introduce a framework, \textbf{K}nowledge-\textbf{E}nhanced with Fine-Grained \textbf{R}einforced Rewards \textbf{M}edical Report Generation (KERM), to tackle the issue. Our approach refines the input to the LVLM by first utilizing MedCLIP for knowledge retrieval, incorporating relevant lesion fact sentences from a curated knowledge corpus. We then introduce a novel purification module to ensure the retrieved knowledge is contextually relevant to the patient's clinical context. Subsequently, we employ fine-grained rewards to guide these models in generating highly supportive and clinically relevant descriptions, ensuring the alignment of model's outputs with desired behaviors. Experimental results on IU-Xray and MIMIC-CXR datasets validate the effectiveness of our approach in mitigating hallucinations and enhancing report quality.
\end{abstract}

\section{Introduction}

Generating radiology reports from medical images represents a critical endeavor within the realm of medical imaging. The task of manually composing such reports by radiologists is not only time-consuming and labor-intensive but also demands a high level of expertise. Consequently, there is a burgeoning interest in methods for automatically generate medical reports for an X-ray, promising solutions that can alleviate these challenges and enhance the overall efficiency of the diagnostic process~\cite{Chen2020GeneratingRR,Li2023DynamicGE,Yang2021KnowledgeMC}.

\begin{figure}
\includegraphics[width=\columnwidth]{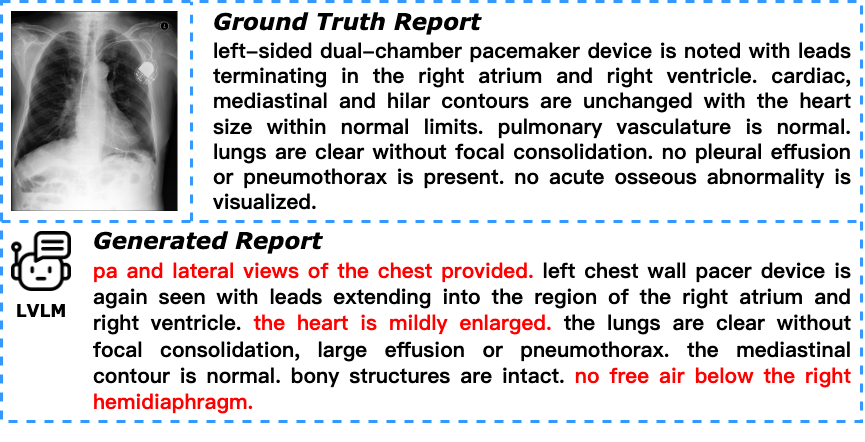}
\caption{An example of the report generated by the LVLM, where the terms marked in red are hallucinations.} \label{example}
\vspace{-0.6cm}
\end{figure}

 The recent advancements in large language models (LLMs)~\cite{touvron2023llama,ouyang2022training} have inspired the development of large vision-language models (LVLMs)~\cite{dai2023instructblip,li2022blip}, which aim to pair these powerful LLMs with image information, building a bridge between the visual and the textual, thus enabling robust comprehension and reasoning across modalities. However, when applying LVLMs to medical report generation, we encountered several challenges, particularly the phenomenon of ``hallucinations'', where the model generates false yet seemingly plausible information. For instance, as illustrated in Figure~\ref{example}, the ground truth report describes a patient ``with a dual-chamber pacemaker'', and the report generated by the LVLM incorrectly suggests ``mild enlargement of the heart'' as well as some extraneous terms, which are not present in the ground truth. Such hallucinations can lead to misdiagnosis and inappropriate treatment plans, with potentially severe consequences for patient care. Prior methods for mitigating LVLMs' hallucinations have focused on refining the training data and adjusting the model architecture~\cite{Liu2023MitigatingHI,Lee2023VolcanoMM}. However, these approaches have not fully addressed the issue, primarily because they neglect the scarcity of high-quality annotations in medical training datasets. The specificity and precision required for medical reports are difficult to achieve without expert knowledge, which can result in model generating incorrect information. This issue stems from the insufficient guidance provided by a lack of accurate and detailed annotations. Moreover, the long-tail problem is prevalent in medical datasets, with common conditions being overrepresented and rare ones underrepresented. This imbalance may cause the model's outputs to deviate from the expected medical findings.

To address these challenges, we propose a new framework, called \textbf{K}nowledge-\textbf{E}nhanced with Fine-Grained \textbf{R}einforced Rewards \textbf{M}edical Report Generation (KERM). It efficiently and substantially enhances the visual grounding of LVLMs beyond pretrained baselines such as LLaVA~\cite{Liu2023VisualIT}, while simultaneously preserving their capability to generate accurate and detailed descriptions. Given a pretrained LVLM (e.g., LLaVA), firstly, we conduct a knowledge corpus, including medical literature and clinical guidelines selected from public datasets such as MIMIC-CXR~\cite{Johnson2019MIMICCXRAD} and CheXpert~\cite{Irvin2019CheXpertAL}, and enhance the model's input by retrieving external knowledge sources through MedCLIP~\cite{Wang2022MedCLIPCL} and introduces a purification module to refine the relevance of retrieved knowledge to the patient's specific clinical context. We provide the necessary external knowledge to ground the LVLM's understanding, thereby improving the accuracy and relevance of the generated reports. Secondly, we employ fine-grained reward modeling by conducting a dual-level assessment to align the model's output with desired behaviors and mitigate the occurrence of hallucinations. At the disease label level, we evaluate the model's output against known medical labels, ensuring that the diagnoses mentioned are consistent with the image content. At the sentence description level, we utilize GPT-3.5 to scrutinize the coherence and plausibility of the generated sentences, penalizing deviations from the expected medical findings, even if they are not outright incorrect. This encourages the model to generate reports that are not only factually accurate but also aligned with the typical patterns observed in medical practice. Experimental results on a public dataset, MIMIC-CXR~\cite{Johnson2019MIMICCXRAD}, confirm the validity and effectiveness of our proposed approach. 
%\revisedhl{(I do not see how the approach addresses challenge ii and iii.)}

Overall, the main contributions of this work are:
\begin{itemize}
\vspace{-0.1cm}
\item[$\bullet$] We introduce a knowledge-enhanced approach, which integrates a curated knowledge corpus sourced from public datasets. It can fortifies the LVLM's input with external knowledge, ensuring that the generated medical reports are grounded in accurate and relevant medical information, thereby enhancing the model's ability to produce reliable and detailed descriptions. 
\vspace{-0.2cm}
\item[$\bullet$]We develop fine-grained reinforced reward modeling that penalizes hallucinatory content from the perspectives of disease-level and sentence-level respectively, promoting outputs that closely align with medical norms and mitigating the occurrence of hallucinations.
\vspace{-0.2cm}
\item[$\bullet$] We conduct comprehensive experiments to demonstrate the effectiveness of our proposed method, which outperforms existing methods on both Natural Language Generation and clinical efficacy metrics.
\end{itemize}

\section{Related Work}
\begin{figure*}[t]
\includegraphics[width=\textwidth]{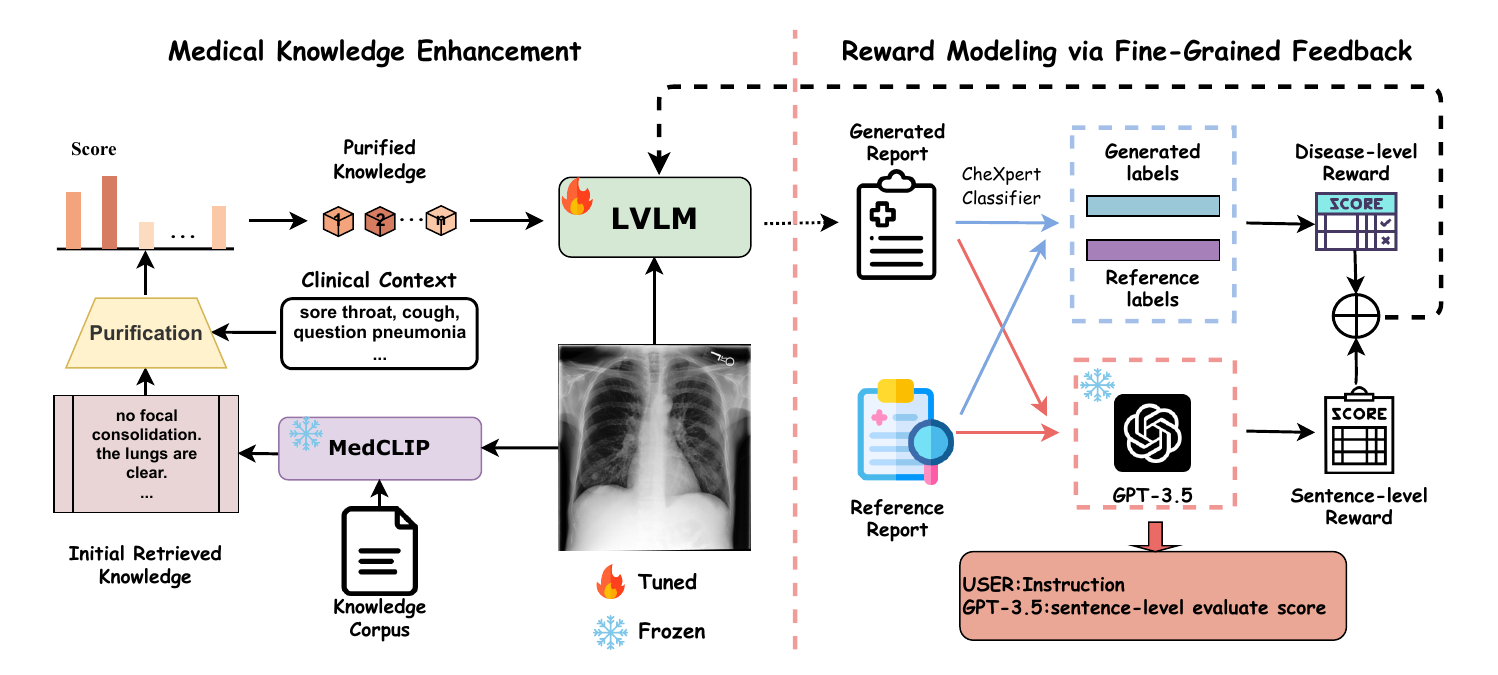}
\caption{Overview of KERM. We first retrieve the knowledge from our constructed Knowledge Corpus to enhance the image representation as additional input. During the training period, we employ CheXpert to obtain disease labels, applying penalties to hallucinatory content at both the disease and sentence levels. This reward is then feedback to the LVLM, thereby guiding the model's performance.} \label{fig1}
% \vspace{-0.4cm}
\end{figure*}

\subsection{Medical Report Generation}
The domain of Medical Report Generation (MRG) in medical artificial intelligence (AI) has surged recently. Early research~\cite{10.1145/3286606.3286863} drew inspiration from image captioning models, using deep Convolutional Neural Networks (CNNs) and Recurrent Neural Networks (RNNs) in an encoder-decoder format~\cite{Vinyals2014ShowAT}.Several studies introduced auxiliary classification tasks to predict medical abnormalities~\cite{Shin2016LearningTR,Wang2018TieNetTE} , enhancing structured guidance for report generation. The attention mechanism improved the integration of visual and linguistic modalities in MRG systems~\cite{Jing2017OnTA,Chen2020GeneratingRR}.

To bridge visual observations and medical domain knowledge, numerous visionand-
language pre-training methods have been devised to
incorporate domain-specific knowledge~\cite{Li2020AuxiliarySK,Li2023DynamicGE}.Generative language modeling evolved from RNNs to transformer architectures, including Large Language Models (LLMs) like LLaMA ~\cite{touvron2023llama}, improving clinical accuracy.
Some studies used reinforcement learning (RL) to optimize clinical relevance~\cite{Liu2019ClinicallyAC,Miura2020ImprovingFC}. However, reliance on models like CheXbert or RadGraph for clinical entity extraction complicates optimization.
\subsection{Large Vision-Language Models}
In recent years, the integration of large language models (LLMs) into multimodal domains has garnered considerable attention~\cite{ouyang2022training,touvron2023llama}. This surge has led to the development of large vision-language models (LVLMs) powered by LLMs~\cite{Ye2023mPLUGOwlME,dai2023instructblip,li2022blip}, enabling comprehension of multimodal inputs and performance of diverse tasks under instructions.

LVLMs typically follow a paradigm where a multimodal alignment module comprehends inputs, followed by a LLM generating responses. For instance, mPLUG-Owl~\cite{Ye2023mPLUGOwlME} pre-trains the encoder and alignment module and finetunes LLaMa~\cite{touvron2023llama} using low-rank adaption. Conversely, LLaVA~\cite{Liu2023VisualIT} pre-trains only the alignment network and finetunes it alongside Vicuna~\cite{Peng2023InstructionTW} based on constructed instructions. MiniGPT-4~\cite{Zhu2023MiniGPT4EV} focuses on finetuning the cross-modal alignment network while freezing other modules. 

Recent advancements also include the development of multimodal biomedical chatbots and generalist models. ELIXR, based on the BLIP-2 framework~\cite{Li2023blip2BL}, trains for contrastive and generative tasks on X-ray image-report pairs, although its evaluation remains private due to the proprietary PaLM-2 model. In contrast, MedPaLM~\cite{Tu2023TowardsGB} proposes a private, PaLM-based generalist model demonstrating impressive performance across various medical tasks and image types, including VQA, image classification, and report generation. However, neither prioritizes the generation and comprehension of X-ray reports, and they appear to lack clinical accuracy, leading to hallucinations, when evaluated for medical image interpretation.

\section{Method}
In this section, we will introduce the detailed implementations of our proposed \textbf{K}nowledge-\textbf{E}nhanced with Fine-Grained \textbf{R}einforced Rewards \textbf{M}edical Report Generation (KERM). We first introduce the overview of our model, then present the proposed modules, Medical Knowledge Enhancement(MKE) and Reward Modeling via Fine-Grained Feedback(RM), respectively.
\subsection{Overview}
The overall architecture of our framework is illustrated in Figure~\ref{fig1}. It's based on a LVLM, composed of a Medical Knowledge Enhancement branch and a Reward Modeling via Fine-Grained Feedback branch. Given an input medical image $I$, the system processes it through a visual encoder to obtain image features $F_{I}$. These features, along with the retrieved knowledge, are then input into the LVLM to generate a descriptive medical report $ R=\{ y_1, y_2, \ldots ,y_{n} \}$, where $y_i$ is a token and $n$ is the length of the report. We formulate our approach as:
\begin{gather}
K_{retrieved} = \mathrm{MKE}(I, C), \\
R = \mathrm{LVLM}((F_I, K_{retrieved})).
\end{gather}
where $\mathrm{MKE}(\cdot)$ represents the Medical Knowledge Enhancement branch. $K_{retrieved}$ stands for the knowledge retrieved by MedCLIP that is most relevant to the image, with $C$ representing the Knowledge Corpus. The final report $R$ is obtained by decoding the internal states of the LVLM, which are influenced by both the image features and the external knowledge.

Given the ground truth report $ R^*=\{ y^*_1, y^*_2, \ldots ,y^*_{n} \}$, we can train the model by minimizing a combined loss function that includes cross-entropy loss for language generation and a reinforcement loss guided by the fine-grained rewards:
\begin{gather}
\mathcal{L}_{RL} =\mathrm{RM}(R, R^*) \\
\mathcal{L}_{\mathrm{CE}}(\theta)= -\sum_{i=1}^{n} \log p_{\theta}(y_i=y^*_i|y^*_{1:i-1},I) \\
\mathcal{L} = \mathcal{L}_{CE} +\mathcal{L}_{RL}
\end{gather}
where $\mathrm{RM}(\cdot)$ denotes the Reward Modeling via Fine-Grained Feedback branch, and $\mathcal{L}_{RL}$ is the reinforcement loss based on the rewards which we will explain in Section~\ref{sec:RL Loss}.

% \vspace{-0.8cm}	
\subsection{Medical Knowledge Enhancement}
 To generate accurate radiology reports from medical images, understanding the medical context and relationships depicted in the images is crucial. This requires not only visual recognition but also the ability to interpret the significance of visual features in relation to medical knowledge. Inspired by~\cite{Li2023KERMKE} , we first construct a medical knowledge corpus and then utilize a pretrained multimodal model MedCLIP~\cite{Wang2022MedCLIPCL} to retrieve relevant facts for each image view, and then apply a purification module to refine the relevance of retrieved knowledge to the patient's specific clinical context. At each step t, the input image with its retrieved  knowledge are fed into the LVLM to ground the model's understanding so as to guide better report generation.
 \subsubsection{Knowledge Corpus Construction}

% \vspace{+0.3cm}
 The knowledge base serves as a repository of medical facts that describe the visual content of medical images. To compile a comprehensive and diverse set of medical descriptions, we parse region descriptions from the medical imaging datasets MIMIC-CXR and CheXpert, focusing on their training sets. After removing duplicates, we construct a knowledge corpus consisting of 100k facts expressed in medical language descriptions, which serve as a Knowledge Corpus for our proposed KERM framework.

\subsubsection{Knowledge Retrieval}
% \vspace{+0.3cm}
 Our objective is to associate each medical image with relevant facts that enhance the model's understanding of the visual content. We employ a pretrained model MedCLIP, which includes an image encoder and a text encoder that map images and text into a shared embedding space. The text encoder is used to encode all facts in the knowledge corpus as search keys, while the image encoder processes the related images as queries. We then identify the facts with the highest cosine similarity scores to the image queries. For each image, we retain the top-10 facts with the highest scores as the initial retrieval knowledge. 

 \subsubsection{Purification Module}
 \label{sec:Purification Module}
 Given the high stakes in medical report generation, it is imperative that the knowledge items selected are not only accurate but also highly pertinent to the patient's clinical narrative, including indications and medical history. Therefore, we propose a purification module in our to distill the most contextually relevant knowledge from the initial top-$k$ retrieval result, ensuring that the retrieved facts are optimally aligned with the patient's specific clinical context. Specially, we construct a context embedding $E_C$ that encapsulates the clinical needs and historical features of the patient derived from their \textit{indications} and \textit{clinical history}. Let $ K=\{ k_1, k_2, \ldots ,k_{t} \}$ represent the initial top-$k$ retrieved facts, each fact $k_i$ is encoded into an embedding $E_{k_i}$ to facilitate the calculation of its similarity to the context vector. Then we computes the cosine similarity between these vectors to quantify the relevance score $s_i$ for each fact, leveraging this score to re-rank the items and prioritize those most contextually aligned with the patient's clinical narrative. The top-5 items, deemed most relevant based on these scores, are selected to form the purified knowledge set $K'$, informing the report generation process.
 
\subsection{Reward Modeling via Fine-Grained Feedback}
In our approach to enhancing the accuracy and coherence of medical report generation, we have developed a novel reinforcement learning strategy that incorporates dual-level reward modeling. This strategy is meticulously designed to mitigate of hallucinations by providing granular feedback at both the disease label and sentence description levels.
\subsubsection{Disease-level Reward}

% \vspace{+0.3cm}
 We employ the CheXPert~\cite{Irvin2019CheXpertAL} labeling tool to label generated reports and the reference reports in 14 different medical terminologies. We calculate the F1 score as the disease-level reward score $\mathbf{R}_{dis}$ for each label to assess the alignment between the model's output and the actual medical findings. The F1 score is a robust measure that balances the trade-off between precision and recall, ensuring that the model's predictions are not only correct but also comprehensive. TP (true positives), FP (false positives), and FN (false negatives) are used to calculate this score, representing correct diagnoses, incorrect diagnoses, and missed diagnoses, respectively.
% The formula for the F1 score is as follows:
% \begin{equation}
% Precision = \frac{TP}{TP + FP} 
% \end{equation}
% \begin{equation}
% Recall = \frac{TP}{TP + FN}
% \end{equation}
% \begin{equation}
% R_{dis} =2\times \frac{Precision×Recall}{Precision +Recall} 
% \end{equation}
% Here, $\mathbf{TP}$ represents the number of true positive predictions (correct diagnoses), $\mathbf{FP}$ is the number of false positive predictions (incorrect diagnoses), and $\mathbf{FN}$ is the number of false negative predictions (missed diagnoses).
\begin{figure}
\includegraphics[width=\columnwidth]{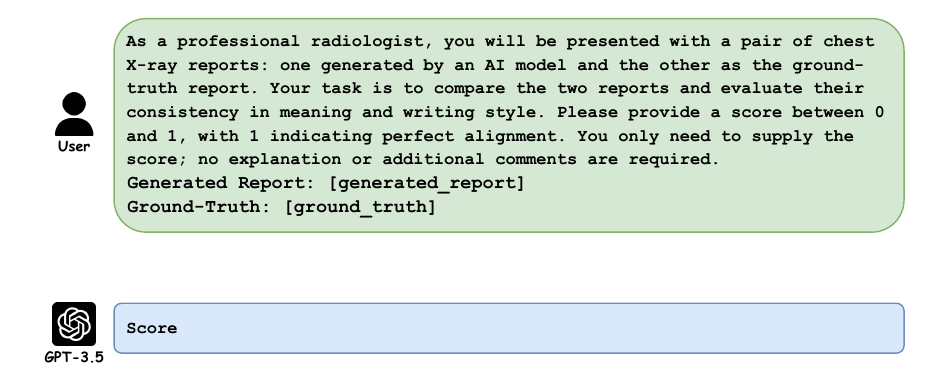}
\caption{ The prompt for generating sentence-level score that scored by GPT-3.5.} \label{prompt}
\vspace{-0.3cm}
\end{figure}
% \subsubsection{Sentence-level Reward}

% \vspace{+0.3cm}
 At the sentence level, we leverage the advanced language understanding capabilities of GPT-3.5 to assess the coherence and plausibility of the generated sentences. We provide GPT-3.5 with sentence pairs, where one is from the generated report and the other from the reference report, along with detailed evaluation instruction as shown in Figure~\ref{prompt}. GPT-3.5 scores the similarity between these pairs ranging from 0 to 1 , with a score closer to 1 indicating a higher degree of coherence and plausibility. This score, $\mathbf{R}_{sen}$, serves as the sentence-level reward.
\subsubsection{Reinforcement Algorithm Loss}
 \label{sec:RL Loss}
% \vspace{+0.3cm}
 Since the decoded text cannot provide gradient information for model training, we harness the Reinforce Algorithm~\cite{Sutton1999PolicyGM} to design a loss function aimed at achieving these goals. At each training step, we sample text sequences from the probability distribution $\mathbf{p}$, which is derived from the softmax function applied to the LVLM's logits. The cumulative reward for each sequence is a weighted blend of $\mathbf{R}_{dis}$ and $\mathbf{R}_{sen}$, with a hyperparameter $\mathbf{\alpha}$ adjusting the emphasis between disease label and sentence description assessments.The loss function of reinforcement algorithm, which incorporates these reward scores, denoted as $\mathcal{L}_{RL}$:
\begin{equation}
R_{t} =\left ( 1-\alpha \right ) R_{dis,t} +  \alpha R_{sen,t}
\end{equation}
\begin{equation}
\mathcal{L}_{RL} = {\textstyle \sum_{t=1}^{T}}p\cdot  R_{t}\cdot log\left ( a_{t}\mid s_{t} \right )
\end{equation}
where $\mathbf{T}$ represents the length of the generated text, $\mathbf{a_{t}}$ is the token sampled at step t, $\mathbf{s_{t}}$ is the corresponding state, $\mathbf{\alpha}$ represents hyperparameter, and $\mathbf{R_{t}}$ represents the reward obtained for the current text.
% Finally, we calculate the sum of cross-entropy loss and reinforcement algorithm loss as our total loss function, is conducted as  $\mathcal{L}$:
% \begin{equation}
% \mathcal{L} = \mathcal{L}_{CE} +\mathcal{L}_{RL}
% \end{equation}
\section{Experiment}
% \subsection{Experiment Settings}
\begin{table*}[t]
% \vspace{-0.4cm}
\centering
	% \scriptsize
    \resizebox{\textwidth}{!}{%
		\begin{tabular}{c|c|cccccc|ccc}
			\hline
			\multirow{2}{*}{Dataset} & \multirow{2}{*}{Model} & \multicolumn{6}{c|}{NLG Metrics} & \multicolumn{3}{c}{CE Metrics}\\ 
            % \cline{2-10} 
		  & & BL-1 & BL-2 & BL-3 & BL-4 & MTR & RG-L & P & R & F1 \\
			\hline
                \multirow{10}{*}{IU-Xray} & HRGR & 0.438 & 0.298 & 0.208 & 0.151 & - & 0.322 & - & - & - \\
			& CoAtt & 0.455 & 0.288 & 0.205 & 0.154 & - & 0.369 & - & - & - \\
			& PKERRG & 0.450 & 0.301 & 0.213 & 0.158 & - & 0.384 & - & - & - \\
			& CMAS-RL & 0.464 & 0.301 & 0.210 & 0.154 & - & 0.362 & - & - & - \\
			& R2Gen & 0.470 & 0.304 & 0.219 & 0.165 & 0.187 & 0.371 & - & - & - \\
			& CMN & 0.475 & 0.309 & 0.222 & 0.170 & 0.191 & 0.375 & - & - & - \\
			& PPKED & 0.483 & 0.315 & 0.224 & 0.168 & 0.190 & 0.376 & - & - & - \\
			& Multicriteria & 0.496 & 0.319 & 0.241 & 0.175 & - & 0.377 & - & - & - \\
			& KM & 0.496 & 0.327 & 0.238 & 0.178 & - & 0.381 & - & - & - \\
			& KERM &  \textbf{0.511} & \textbf{0.333} & \textbf{0.249} & \textbf{0.182} & \textbf{0.197} & \textbf{0.388} & - & - & - \\
			\hline
			\multirow{7}{*}{MIMIC-CXR} & CCR & 0.313 & 0.206 & 0.146 & 0.103 & - & \textbf{0.306} & - & - & - \\
			& Multicriteria & 0.351 & 0.223 & 0.157 & \textbf{0.118} & - & 0.287 & - & - & - \\
			& R2Gen & 0.353 & 0.218 & 0.145 & 0.103 & 0.142 & 0.277 & 0.333 & 0.273 & 0.276 \\
			& CMN & 0.353 & 0.218 & 0.148 & 0.106 & 0.142 & 0.278 & 0.334 & 0.275 & 0.278 \\
			& PPKED & 0.360 & 0.224 & 0.149 & 0.106 & 0.149 & 0.284 & - & - & - \\
			& KM & 0.363 & 0.228 & 0.156 & 0.115 & - & 0.284 & \textbf{0.458} & 0.348 & 0.371 \\
			& KERM &\textbf{ 0.378} & \textbf{0.235} & \textbf{0.157} & 0.109 & \textbf{0.152} & 0.283 & 0.394 & \textbf{0.436} & \textbf{0.415} \\
			\hline
		\end{tabular}
    }
\caption{Comparisons of our model with previous studies on the IU X-Ray and MIMIC-CXR test set with respect to natural language generation (NLG) and  clinical efficacy (CE) metrics. BL-n denotes BLEU score using up to n-grams; MTR and RG-L denote METEOR and ROUGE-L, respectively. P, R and F1 represent precision, recall and F1-score, respectively. KERM is our proposed model. Best results are in bold.}
\label{table:comparisons_with_previous}
% \vspace{-0.4cm}	
\end{table*}
\subsection{Dataset}
We evaluate our proposed KERM on two widely-used radiology reporting benchmark, IU-Xray~\cite{DemnerFushman2015PreparingAC} and MIMIC-CXR~\cite{Johnson2019MIMICCXRAD}, to verify the model’s effectiveness. To ensure a fair comparison, we adopt the settings in ~\cite{Chen2020GeneratingRR} for report preprocessing. 

\textbf{IU-Xray} is a publicly available radiological dataset collected by Indiana University, with 7,470 frontal and lateral-view chest X-ray images and 3,955 reports. The reports include \textit{impression}, \textit{findings}, \textit{comparison}, and \textit{indication} sections. Following \cite{Li2018HybridRR}, we excluded images without reports and there are 5,910 images and 2,955 reports left for this study. Following \cite{Chen2020GeneratingRR}, we split the data into training/validation/test set by 7:1:2 of the dataset, and took the \textit{impression} and the \textit{findings} sections as the target captions to be generated.

\textbf{MIMIC-CXR} is the largest radiology image dataset so far, sourcing from the Beth Israel Deaconess Medical Center between 2011-2016. We followed ~\cite{Liu2021ExploringAD} to adopt an alpha version of 473, 057 Chest X-ray images and 206, 563 reports from 63, 478 patients. Each study comprises multiple sections, including \textit{comparison}, \textit{clinical history}, \textit{indication}, \textit{reasons for examination}, \textit{impressions}, and \textit{findings}. We adopted the official split of training/validation/test set, and took the \textit{findings} section as the target captions to be generated.

\subsection{Baselines and Evaluation Metrics}

% \vspace{+0.3cm}
\noindent \textbf{Baselines} we compare our KERM with a wide range of existing state-of-the-art MRG systems on the benchmark, including R2Gen~\cite{Chen2020GeneratingRR}, HRGR~\cite{Li2018HybridRR}, CoAtt~\cite{Jing2017OnTA}, PKERRG~\cite{Wang2022PriorKE}, CMAS-RL~\cite{Jing2019ShowDA}, CMN~\cite{Chen2022CrossmodalMN}, CCR~\cite{Liu2019ClinicallyAC}, PPKED~\cite{Liu2021ExploringAD}, KM~\cite{Yang2021KnowledgeMC} and Multicriteria~\cite{Wang2022AutomatedRR} . Since we follow the same settings, we directly cite the results from original papers.
\vspace{+0.3cm}

\noindent \textbf{Evaluation Metrics} We utilize automatic Natural Language Generation (NLG) evaluation metrics such as CIDEr~\cite{Vedantam2014CIDErCI}, ROUGE-L~\cite{Lin2004ROUGEAP}, and BLEU~\cite{Papineni2002BleuAM}, which quantify the correlation between two text sequences statistically. However, these metrics, which are limited to n-grams of up to 4, may not fully capture the nuances of disease states due to the prevalence of negations in medical language, where negation cues and disease terms can be spatially distant within a sentence. To address this, we incorporate medical abnormality detection as an additional metric.
Specifically, we assess the generated reports against the ground truth by comparing the CheXpert~\cite{Irvin2019CheXpertAL} labeled annotations for certain categories within the 14 diseases. For this comparison, we calculate the F1-Score, precision, and recall for all models, ensuring a comprehensive evaluation of their performance.
% \vspace{-0.4cm} 
\subsection{Implementation Details}
 In our experiments, we adopt the pretrained MedCLIP\cite{Wang2022MedCLIPCL} to retrieve facts for each image. And we employ the LVLM, LLaVA-1.5-7b~\cite{Liu2023VisualIT} as the backbone, and then we employ LoRA-tuning~\cite{Hu2021LoRALA} and deepspeed zero stage 3 to conduct minimal training on the model for 1 epoch. The learning rate is set as 2e-4 and the optimizer is AdamW~\cite{Loshchilov2017DecoupledWD} with a weight decay of 0.02. During the training phase, we initiate a warm-up ratio of 0.03, after which we apply the cosine schedule to decay the learning rate. We set $\mathbf{\alpha}$ to 0.4, based on a hyperparameter search . All of the experiments are conducted on 8 NVIDIA GeForce RTX3090 GPUs.

\subsection{Results and Discussion}
\subsubsection{Main Results}
% \noindent \textbf{Main Results}
% \vspace{-0.4cm}	

Table~\ref{table:comparisons_with_previous} presents the comparison results across both Natural Language Generation (NLG) and clinical efficacy (CE) metrics on both MIMIC-CXR and IU X-Ray. On IU X-Ray, our method significantly outperforms methods in previous studies in all NLG metrics. Specifically, KERM achieves BL-4 score of 0.182, MTR score of 0.197, and RG-L score of 0.388. This demonstrates that our model excels not only in generating accurate words and phrases but also in constructing coherent long sentences and maintaining logical flow between sentences. On MIMIC-CXR, it is observed that our method surpasses existing methods in most NLG metrics and achieves comparable performance to the state-of-the-art in BL-4 and MTR. This indicates a robust capability in capturing the nuances of medical language and adhering to clinical standards. The RG-L metric may not be optimal because the order of lesions or sentences in the reports generated by our model does not strictly align with the ground-truth order. In the three CE metrics, our method significantly outperforms previous methods, which indicates that
our model predicts much fewer false positive and false negative diseases, respectively. Although our method has a lower precision compared to the KM method, it exceeds KM in the more comprehensive F1-score metric. The significant improvements in CE metrics are a direct result of our approach, which enriches the model's understanding by retrieving factual knowledge from a comprehensive corpus. This is complemented by a fine-grained reward model that penalizes inaccuracies and deviations, ensuring the generation of contextually appropriate and clinically sound reports.

% \subsubsection{Hyperparameter Analysis}
\begin{table}[ht]
\resizebox{\columnwidth}{!}{
\begin{tabular}{c|cccccc}
\hline
Settings & BL-1 & BL-2 & BL-3 & BL-4 & MTR & RG-L \\
\hline
Base & 0.445 & 0.295 & 0.210 & 0.162 & 0.180 & 0.372 \\
w/MKE & 0.475 & 0.308 & 0.222 & 0.170 & 0.191 & 0.385 \\
w/RM & 0.455 & 0.302 & 0.217 & 0.165 & 0.187 & 0.380 \\
\textbf{KERM} & \textbf{0.511} & \textbf{0.333} & \textbf{0.249} & \textbf{0.182} & \textbf{0.197} & \textbf{0.388} \\
\hline
\end{tabular}}
\caption{The comparison of natural language generation (NLG) metrics on IU X-Ray dataset. ``$w/(\cdot)$'' means the application of the module.}
\label{tab:ablation_study_iu}
\end{table}
% \vspace{-0.8cm} 
\begin{table*}[h]
\centering
\resizebox{\textwidth}{!}{
\begin{tabular}{c|ccc|cccccc|ccc} 
\toprule
Settings & MKE  & $R_{dis}$  & $R_{sen}$ &  BL-1 & BL-2 & BL-3 & BL-4 & MTR & RG-L & P & R & F1  \\ 
\midrule
% \midrule
Base     & \XSolidBrush& \XSolidBrush &  \XSolidBrush & 0.337 & 0.203 & 0.132 & 0.098& 0.131 & 0.273 & 0.296 & 0.163 & 0.153 \\ 
\midrule
(a)      & \checkmark &  \XSolidBrush &  \XSolidBrush & 0.361 & 0.222 & 0.149 & 0.103 & 0.142 & 0.278 & 0.332 & 0.264 &  0.297   \\
(b)      &   \XSolidBrush  & \checkmark & \checkmark & 0.352  & 0.216 & 0.144 & 0.101 & 0.135 & 0.275 & 0.322 & 0.253 & 0.282  \\ 
\midrule
(c)      & \checkmark &\XSolidBrush& \checkmark & 0.370 & 0.231 & 0.154 & \textbf{0.112} & 0.145 & \textbf{0.285} & 0.359 & 0.280 & 0.315  \\
(d)      & \checkmark & \checkmark & \XSolidBrush & 0.368& 0.223 & 0.145 & 0.106 & 0.141 & 0.279 & 0.363 & 0.282 & 0.317 \\ 
\midrule
KERM      & \checkmark & \checkmark & \checkmark &\textbf{ 0.378} & \textbf{0.235} & \textbf{0.157} & 0.109 & \textbf{0.152} & 0.283 & \textbf{0.394} & \textbf{0.436} & \textbf{0.415}  \\
\bottomrule
\end{tabular}}
\caption{Quantitative analysis of proposed method on MIMIC-CXR dataset. MKE, $R_{dis}$ and $R_{sen}$ represent Medical Knowledge Enhancement, disease-level and sentence-level feedback, respectively.}
\label{tab:ablationstudy}
% \vspace{-0.4cm}
\end{table*}

\subsubsection{Ablation study}
In this section, we conduct ablation studies on IU-Xray and MIMIC-CXR datasets to investigate the contribution of each component in our proposed KERM. Table~\ref{tab:ablationstudy} presents the quantitative analysis of KERM on MIMIC-CXR across both NLG and CE metrics. And cmeasuring descriptive accuracy is reported in Table~\ref{tab:ablation_study_iu}. Our base model is LLaVA-1.5-7b.
\begin{figure}[ht]
\centering
\includegraphics[width=\columnwidth]{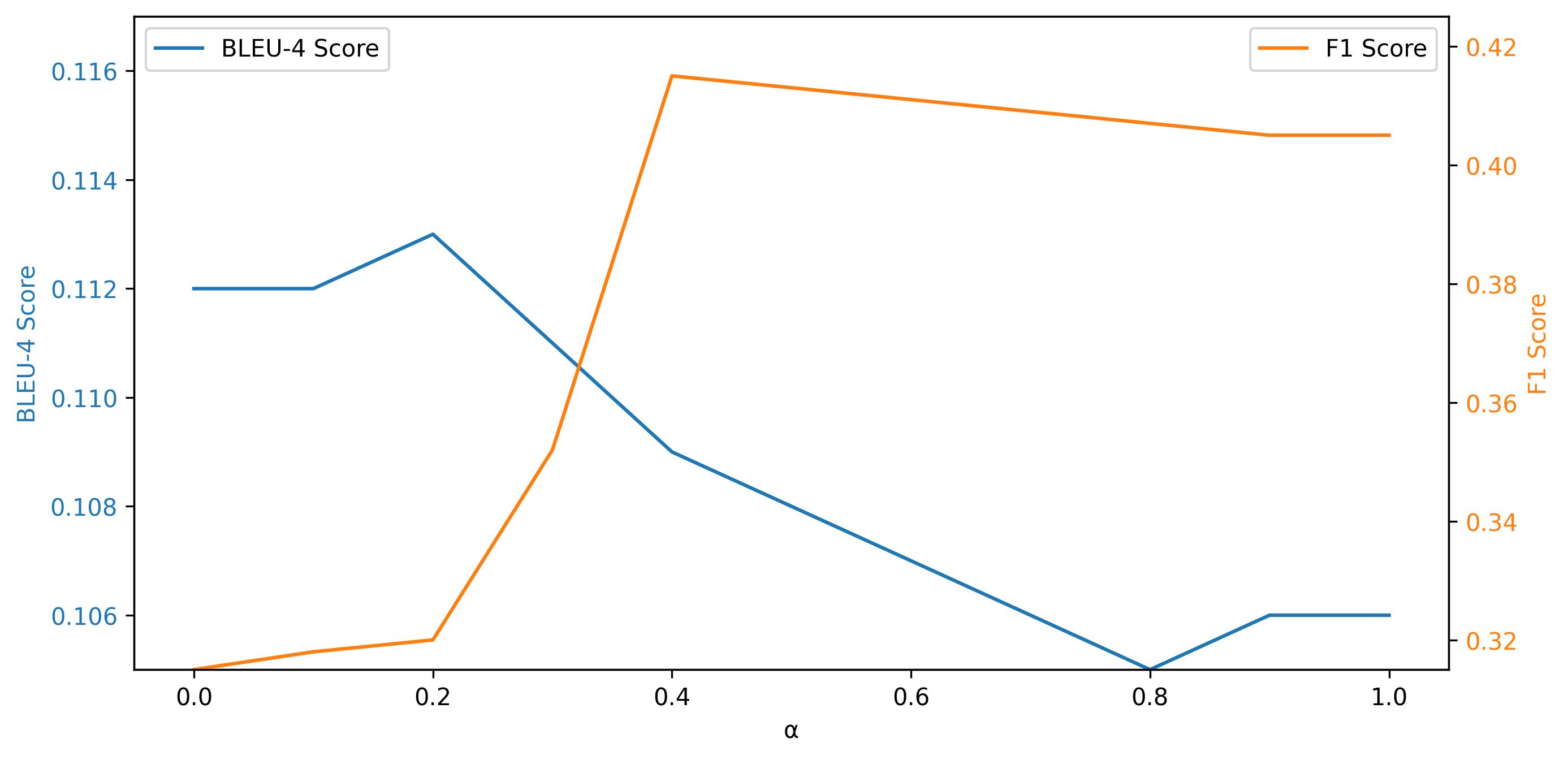}
\caption{Analysis of the hyperparameter $\mathbf{\alpha}$ with respect to F1 and BLEU-4 on MIMIC-CXR dataset.} \label{Hyperparameter}
% \vspace{-0.5cm}
\end{figure}

\noindent \textbf{Effect of The Components and Submodules} It can be observed that adding MKE(Medical Knowledge Enhancement) and RM(Reward Modeling via Fine-Grained Feedback) on both the MIMIC-CXR and IU X-Ray datasets individually, in comparison to the baseline model, leads to significant improvements on all metrics. This observation indicates the effectiveness of both modules. MKE exhibits greater enhancement compared to RM. This might stem from the fact that the knowledge, obtained through retrieval, are more closely related to the current image. These knowledge contain additional detailed information, such as position and existence. Incorporating fine-grained rewards shows substantial growth, with the introduction of reward scores effectively mitigating the issue of hallucinations. This encourages the model to focus on avoiding inaccuracies and deviations. 

Furthermore, comparing (c) and (d) in Table~\ref{tab:ablationstudy}, it is observed that $R_{dis}$ brings more improvement than $R_{sen}$ on the NLG metrics, while the opposite is true on the CE metrics. We speculate the reason is that disease-level reward can more effectively improve the model to identify the existence of diseases and sentence-level reward promotes outputs that closely align with medical norms. Ultimately, the integration of such three improvements yields the best overall performance. 

Ultimately, the integration of MKE and RM, as seen in the KERM model, yields the best overall performance on both datasets. This synergistic effect results in highly accurate and clinically relevant medical reports, reflecting the model's enhanced diagnostic capabilities and the reliability of its generated radiology reports.

\begin{figure*}[ht]
\centering
\includegraphics[width=\textwidth]{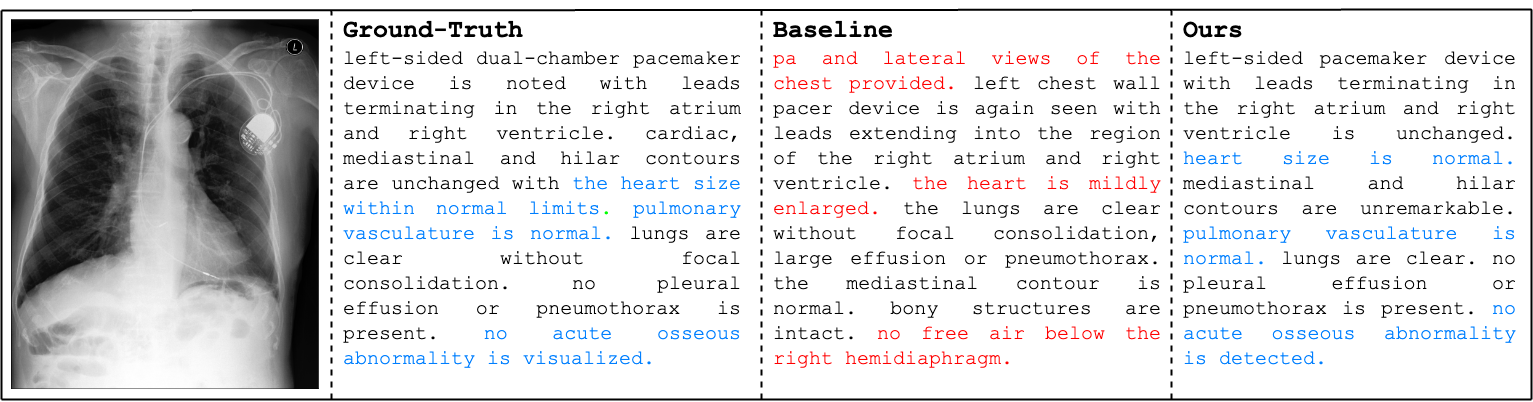}
\caption{Illustrations of reports from ground truth, ours and Base. For better visualization, different colors highlight different medical terms. The terms marked in red are hallucinations, the terms marked in blue means descriptions included in Ground-Truth but not mentioned in the base model.} \label{casestudy}
\end{figure*} 
% \vspace{+0.2cm}
\noindent\textbf{Hyperparameter Analysis} We also conduct an ablation study on the hyperparameter $\mathbf{\alpha}$ to investigate at which value can better enhance the model’s performence of generating accurate and consistent report on MIMIC-CXR dataset. As is shown in Figure~\ref{Hyperparameter}, $\mathbf{\alpha}$ is analyzed with values ranging from 0 to 1 in terms of F1 and BLEU-4 scores. Overall, the performance remains stable across a wide range of $\mathbf{\alpha}$, as the fluctuations of F1 and BLEU-4 are within 10\% and 1.2\%, respectively. $\mathbf{\alpha}=0.4$ performs better in F1 and BLEU-4 scores, which is the value we used in the experiments.
% \vspace{-0.8cm}	

\subsubsection{Case Study}
% \noindent\textbf{Case Study}
% \vspace{-0.4cm}
To further investigate the effectiveness of our method, we provide a qualitative comparison to the base model (LVLM) in Figure~\ref{casestudy}, where different colors on the texts indicate different medical terms(more cases can be seen in Appendix~\ref{sec:More Cases}). It is observed that our model generates descriptions that closely align with the ground-truth report in terms of content flow. Furthermore, as shown in Figure~\ref{casestudy}, we have found that KERM covers almost all of the necessary medical terms and abnormalities in the ground-truth reports, this comprehensive coverage is a significant improvement over the base model, which often misses crucial medical details. The performance of KERM proves that the reports generated from our model are comprehensive and accurate compared to the base model, effectively alleviating hallucinations.

\section{Conclusions and Future Work }
In this paper, we introduce KERM, a new framework designed to enhance the accuracy and reliability of radiology report generation from medical images. KERM addresses the critical challenge of hallucinations in the LVLM by retrieving fact knowledge from a comprehensive corpus and introducing a purification module to ensure contextual relevance, which enriches the model's understanding. This approach is complemented by fine-grained reward modeling, which penalizes both disease-level inaccuracies and sentence-level deviations from the expected medical findings. Our method's effectiveness is validated through extensive experiments, showcasing its potential to significantly improve the diagnostic process. In the future, we plan to develop more comprehensive evaluation metrics to better assess hallucinations in medical reports.

\section{Limitations}
% Our study, while showcasing the efficacy of the KERM framework in enhancing medical report generation, is not without limitations. The framework's performance is contingent upon the comprehensiveness of the knowledge corpus and the accuracy of the Purification module in aligning with patient-specific contexts. Additionally, the fine-grained reward system, though effective, may not encapsulate the entirety of clinical nuances. The generalizability of our results to other medical domains and datasets remains to be validated. Finally, the computational demands of our approach could be prohibitive for some settings. Future iterations of KERM will aim to address these constraints and expand its applicability.
While our KERM framework has demonstrated significant improvements in the accuracy and reliability of medical report generation, there are several limitations that warrant discussion. Firstly, the performance of KERM is inherently dependent on the quality and comprehensiveness of the knowledge corpus used for knowledge retrieval. Should the corpus lack certain medical facts or contain outdated information, it could potentially lead to omissions or inaccuracies in the generated reports.

Secondly, the Purification module, although designed to enhance the contextual relevance of the retrieved knowledge, may not always perfectly align with the specific nuances of each patient's clinical narrative. This could be due to the complexity of medical cases and the variability in how clinical history is documented.

Additionally, our framework's reliance on fine-grained rewards for guiding the generation process assumes that the reward model accurately reflects all aspects of clinical relevance and accuracy. However, the model's ability to capture the full spectrum of medical knowledge and the subtleties of medical language is subject to the training data and the design of the reward system.

Moreover, while our experiments on IU-Xray and MIMIC-CXR datasets have shown promising results, the external validity of our approach may be limited. The generalizability of KERM to other datasets or different medical domains requires further investigation, as the model's performance could vary with changes in data distribution or clinical presentation.

Lastly, the computational expense associated with training and deploying large vision language models like those used in KERM cannot be overlooked. The resource-intensive nature of our approach may pose challenges for implementation in settings with limited computational resources.

In future work, we aim to address these limitations by expanding the knowledge corpus, refining the Purification module, enhancing the reward modeling, and conducting additional experiments across diverse datasets to ensure broader applicability and robustness of our framework.
% \section*{Acknowledgments}
% This research is supported by the National Natural Science Foundation of China (No.62476127),  the Natural Science Foundation of Jiangsu Province (No.BK20242039), the Basic Research Program of the Bureau of Science and Technology (ILF24001), the Meituan Research Fund (No.PO250624101698), the Scientific Research Starting Foundation of Nanjing University of Aeronautics and Astronautics (No.YQR21022), and the High Performance Computing Platform of Nanjing University of Aeronautics and Astronautics.

\section{Ethics Considerations}

The development and application of our KERM framework are grounded in a commitment to ethical standards, particularly concerning the handling of sensitive medical data. Our work strictly adheres to the deidentification protocols and usage policies associated with the IU X-Xray and MIMIC-CXR dataset, ensuring that all patient information remains confidential and is used solely for research purposes.

A critical aspect of our ethical considerations involves the responsible use of large language models (LLMs), such as the gpt-3.5-turbo model deployed on the Azure OpenAI platform. We acknowledge the financial implications of utilizing cloud-based services, recognizing that the cost per thousand tokens can create barriers to access and scalability, potentially limiting the equitable use of advanced AI in medical applications.

Moreover, we are vigilant about the risks associated with LLMs, including the potential for "hallucinations"—the generation of false or misleading information. In the context of medical report generation, where accuracy is paramount, we have implemented strategies to minimize these risks. Our approach prompts the LLM to rephrase existing medical content into coherent and stylistically consistent prose, rather than creating new medical content. This method is designed to leverage the strengths of LLMs in language generation while reducing the likelihood of introducing inaccuracies.

% To further mitigate the risks of hallucination, we provide serialized content to the model as in-context examples, serving as a style injection step that is deliberately separated from the content generation phase. This separation aims to limit the LLM's exposure to incomplete or unverified information, thereby enhancing the reliability of the generated medical reports.

% Our framework also incorporates a purification module to ensure that the knowledge retrieved is not only accurate but also highly pertinent to the patient's clinical narrative. This additional layer of scrutiny is intended to align the model's outputs with the specific clinical context, thus upholding the highest standards of medical reporting.

In conclusion, our ethical considerations are integral to the design and implementation of the KERM framework. We remain dedicated to the responsible use of AI in medicine, prioritizing accuracy, patient confidentiality, and the avoidance of misinformation in medical report generation.

\section{Acknowledgment}
This research is supported by the National Natural Science Foundation of China (No.62476127), the Natural Science Foundation of Jiangsu Province (No.BK20242039), the Scientific Research Starting Foundation of Nanjing University of Aeronautics and Astronautics (No.YQR21022), and the High Performance Computing Platform of Nanjing University of Aeronautics and Astronautics.

% This document has been adapted
% by Steven Bethard, Ryan Cotterell and Rui Yan
% from the instructions for earlier ACL and NAACL proceedings, including those for
% ACL 2019 by Douwe Kiela and Ivan Vuli\'{c},
% NAACL 2019 by Stephanie Lukin and Alla Roskovskaya,
% ACL 2018 by Shay Cohen, Kevin Gimpel, and Wei Lu,
% NAACL 2018 by Margaret Mitchell and Stephanie Lukin,
% Bib\TeX{} suggestions for (NA)ACL 2017/2018 from Jason Eisner,
% ACL 2017 by Dan Gildea and Min-Yen Kan,
% NAACL 2017 by Margaret Mitchell,
% ACL 2012 by Maggie Li and Michael White,
% ACL 2010 by Jing-Shin Chang and Philipp Koehn,
% ACL 2008 by Johanna D. Moore, Simone Teufel, James Allan, and Sadaoki Furui,
% ACL 2005 by Hwee Tou Ng and Kemal Oflazer,
% ACL 2002 by Eugene Charniak and Dekang Lin,
% and earlier ACL and EACL formats written by several people, including
% John Chen, Henry S. Thompson and Donald Walker.
% Additional elements were taken from the formatting instructions of the \emph{International Joint Conference on Artificial Intelligence} and the \emph{Conference on Computer Vision and Pattern Recognition}.

\bibliography{custom}

@inproceedings{10.1145/3286606.3286863,
author = {Allaouzi, Imane and Ben Ahmed, M. and Benamrou, B. and Ouardouz, M.},
title = {Automatic Caption Generation for Medical Images},
year = {2018},
isbn = {9781450365628},
publisher = {Association for Computing Machinery},
address = {New York, NY, USA},
url = {https://doi.org/10.1145/3286606.3286863},
doi = {10.1145/3286606.3286863},
booktitle = {Proceedings of the 3rd International Conference on Smart City Applications},
articleno = {86},
numpages = {6},
location = {Tetouan, Morocco},
series = {SCA '18}
}

@inproceedings{
dai2023instructblip,
title={Instruct{BLIP}: Towards General-purpose Vision-Language Models with Instruction Tuning},
author={Wenliang Dai and Junnan Li and Dongxu Li and Anthony Tiong and Junqi Zhao and Weisheng Wang and Boyang Li and Pascale Fung and Steven Hoi},
booktitle={Thirty-seventh Conference on Neural Information Processing Systems},
year={2023},
url={https://openreview.net/forum?id=vvoWPYqZJA}
}

@article{Vinyals2014ShowAT,
  title={Show and tell: A neural image caption generator},
  author={Oriol Vinyals and Alexander Toshev and Samy Bengio and D. Erhan},
  journal={2015 IEEE Conference on Computer Vision and Pattern Recognition (CVPR)},
  year={2014},
  pages={3156-3164},
  url={https://api.semanticscholar.org/CorpusID:1169492}
}

@article{Yang2021KnowledgeMC,
  title={Knowledge matters: Chest radiology report generation with general and specific knowledge},
  author={Shuxin Yang and Xian Wu and Shen Ge and S.kevin Zhou and Li Xiao},
  journal={Medical image analysis},
  year={2021},
  volume={80},
  pages={
          102510
        },
  url={https://api.semanticscholar.org/CorpusID:249557147}
}

@article{Vedantam2014CIDErCI,
  title={CIDEr: Consensus-based image description evaluation},
  author={Ramakrishna Vedantam and C. Lawrence Zitnick and Devi Parikh},
  journal={2015 IEEE Conference on Computer Vision and Pattern Recognition (CVPR)},
  year={2014},
  pages={4566-4575},
  url={https://api.semanticscholar.org/CorpusID:9026666}
}

@article{ouyang2022training,
  title={Training language models to follow instructions with human feedback},
  author={Ouyang, Long and Wu, Jeffrey and Jiang, Xu and Almeida, Diogo and Wainwright, Carroll and Mishkin, Pamela and Zhang, Chong and Agarwal, Sandhini and Slama, Katarina and Ray, Alex and others},
  journal={Advances in Neural Information Processing Systems},
  volume={35},
  pages={27730--27744},
  year={2022}
}

@article{Li2023DynamicGE,
  title={Dynamic Graph Enhanced Contrastive Learning for Chest X-Ray Report Generation},
  author={Mingjie Li and Bingqian Lin and Zicong Chen and Haokun Lin and Xiaodan Liang and Xiaojun Chang},
  journal={2023 IEEE/CVF Conference on Computer Vision and Pattern Recognition (CVPR)},
  year={2023},
  pages={3334-3343},
  url={https://api.semanticscholar.org/CorpusID:257631847}
}

@inproceedings{Lin2004ROUGEAP,
  title={ROUGE: A Package for Automatic Evaluation of Summaries},
  author={Chin-Yew Lin},
  booktitle={Annual Meeting of the Association for Computational Linguistics},
  year={2004},
  url={https://api.semanticscholar.org/CorpusID:964287}
}

@article{Chen2022CrossmodalMN,
  title={Cross-modal Memory Networks for Radiology Report Generation},
  author={Zhihong Chen and Yaling Shen and Yan Song and Xiang Wan},
  journal={ArXiv},
  year={2022},
  volume={abs/2204.13258},
  url={https://api.semanticscholar.org/CorpusID:236460168}
}

@article{Miura2020ImprovingFC,
  title={Improving Factual Completeness and Consistency of Image-to-Text Radiology Report Generation},
  author={Yasuhide Miura and Yuhao Zhang and C. Langlotz and Dan Jurafsky},
  journal={ArXiv},
  year={2020},
  volume={abs/2010.10042},
  url={https://api.semanticscholar.org/CorpusID:224803298}
}

@inproceedings{Loshchilov2017DecoupledWD,
  title={Decoupled Weight Decay Regularization},
  author={Ilya Loshchilov and Frank Hutter},
  booktitle={International Conference on Learning Representations},
  year={2017},
  url={https://api.semanticscholar.org/CorpusID:53592270}
}

@article{DemnerFushman2015PreparingAC,
  title={Preparing a collection of radiology examinations for distribution and retrieval},
  author={Dina Demner-Fushman and Marc D. Kohli and Marc B. Rosenman and Sonya E. Shooshan and Laritza M. Rodriguez and Sameer Kiran Antani and George R. Thoma and Clement J. McDonald},
  journal={Journal of the American Medical Informatics Association : JAMIA},
  year={2015},
  volume={23 2},
  pages={
          304-10
        },
  url={https://api.semanticscholar.org/CorpusID:16941525}
}

@article{Wang2022PriorKE,
  title={Prior Knowledge Enhances Radiology Report Generation},
  author={Song Wang and Liyan Tang and Mingquan Lin and George L. Shih and Ying Ding and Yifan Peng},
  journal={AMIA ... Annual Symposium proceedings. AMIA Symposium},
  year={2022},
  volume={2022},
  pages={
          486-495
        },
  url={https://api.semanticscholar.org/CorpusID:245853976}
}

@article{Liu2021ExploringAD,
  title={Exploring and Distilling Posterior and Prior Knowledge for Radiology Report Generation},
  author={Fenglin Liu and Xian Wu and Shen Ge and Wei Fan and Yuexian Zou},
  journal={2021 IEEE/CVF Conference on Computer Vision and Pattern Recognition (CVPR)},
  year={2021},
  pages={13748-13757},
  url={https://api.semanticscholar.org/CorpusID:235421693}
}

@article{Hu2021LoRALA,
  title={LoRA: Low-Rank Adaptation of Large Language Models},
  author={J. Edward Hu and Yelong Shen and Phillip Wallis and Zeyuan Allen-Zhu and Yuanzhi Li and Shean Wang and Weizhu Chen},
  journal={ArXiv},
  year={2021},
  volume={abs/2106.09685},
  url={https://api.semanticscholar.org/CorpusID:235458009}
}

@inproceedings{Jing2017OnTA,
  title={On the Automatic Generation of Medical Imaging Reports},
  author={Baoyu Jing and Pengtao Xie and Eric P. Xing},
  booktitle={Annual Meeting of the Association for Computational Linguistics},
  year={2017},
  url={https://api.semanticscholar.org/CorpusID:5776384}
}

@inproceedings{Papineni2002BleuAM,
  title={Bleu: a Method for Automatic Evaluation of Machine Translation},
  author={Kishore Papineni and Salim Roukos and Todd Ward and Wei-Jing Zhu},
  booktitle={Annual Meeting of the Association for Computational Linguistics},
  year={2002},
  url={https://api.semanticscholar.org/CorpusID:11080756}
}

@inproceedings{Jing2019ShowDA,
  title={Show, Describe and Conclude: On Exploiting the Structure Information of Chest X-ray Reports},
  author={Baoyu Jing and Zeya Wang and Eric P. Xing},
  booktitle={Annual Meeting of the Association for Computational Linguistics},
  year={2019},
  url={https://api.semanticscholar.org/CorpusID:196199713}
}

@article{Li2018HybridRR,
  title={Hybrid Retrieval-Generation Reinforced Agent for Medical Image Report Generation},
  author={Yuan Li and Xiaodan Liang and Zhiting Hu and Eric P. Xing},
  journal={ArXiv},
  year={2018},
  volume={abs/1805.08298},
  url={https://api.semanticscholar.org/CorpusID:44234294}
}

@article{Li2023KERMKE,
  title={KERM: Knowledge Enhanced Reasoning for Vision-and-Language Navigation},
  author={Xiangyang Li and Zihan Wang and Jiahao Yang and Yaowei Wang and Shuqiang Jiang},
  journal={2023 IEEE/CVF Conference on Computer Vision and Pattern Recognition (CVPR)},
  year={2023},
  pages={2583-2592},
  url={https://api.semanticscholar.org/CorpusID:257771855}
}

@article{Shin2016LearningTR,
  title={Learning to Read Chest X-Rays: Recurrent Neural Cascade Model for Automated Image Annotation},
  author={Hoo-Chang Shin and Kirk Roberts and Le Lu and Dina Demner-Fushman and Jianhua Yao and Ronald M. Summers},
  journal={2016 IEEE Conference on Computer Vision and Pattern Recognition (CVPR)},
  year={2016},
  pages={2497-2506},
  url={https://api.semanticscholar.org/CorpusID:2750623}
}

@inproceedings{Wang2022MedCLIPCL,
  title={MedCLIP: Contrastive Learning from Unpaired Medical Images and Text},
  author={Zifeng Wang and Zhenbang Wu and Dinesh Agarwal and Jimeng Sun},
  booktitle={Conference on Empirical Methods in Natural Language Processing},
  year={2022},
  url={https://api.semanticscholar.org/CorpusID:252992913}
}

@article{Tu2023TowardsGB,
  title={Towards Generalist Biomedical AI},
  author={Tao Tu and Shekoofeh Azizi and Danny Driess and Mike Schaekermann and Mohamed Amin and Pi-Chuan Chang and Andrew Carroll and Chuck Lau and Ryutaro Tanno and Ira Ktena and Basil Mustafa and Aakanksha Chowdhery and Yun Liu and Simon Kornblith and David J. Fleet and P. A. Mansfield and Sushant Prakash and Renee C Wong and Sunny Virmani and Christopher Semturs and Seyedeh Sara Mahdavi and Bradley Green and Ewa Dominowska and Blaise Ag{\"u}era y Arcas and Jo{\"e}lle K. Barral and Dale R. Webster and Greg S Corrado and Yossi Matias and K. Singhal and Peter R. Florence and Alan Karthikesalingam and Vivek Natarajan},
  journal={ArXiv},
  year={2023},
  volume={abs/2307.14334},
  url={https://api.semanticscholar.org/CorpusID:260164663}
}

@article{Peng2023InstructionTW,
  title={Instruction Tuning with GPT-4},
  author={Baolin Peng and Chunyuan Li and Pengcheng He and Michel Galley and Jianfeng Gao},
  journal={ArXiv},
  year={2023},
  volume={abs/2304.03277},
  url={https://api.semanticscholar.org/CorpusID:257985497}
}

@article{Johnson2019MIMICCXRAD,
  title={MIMIC-CXR, a de-identified publicly available database of chest radiographs with free-text reports},
  author={Alistair E. W. Johnson and Tom J. Pollard and Seth J. Berkowitz and Nathaniel R. Greenbaum and Matthew P. Lungren and Chih-ying Deng and Roger G. Mark and Steven Horng},
  journal={Scientific Data},
  year={2019},
  volume={6},
  url={https://api.semanticscholar.org/CorpusID:209342303}
}

@article{Lee2023VolcanoMM,
  title={Volcano: Mitigating Multimodal Hallucination through Self-Feedback Guided Revision},
  author={Seongyun Lee and Sue Hyun Park and Yongrae Jo and Minjoon Seo},
  journal={ArXiv},
  year={2023},
  volume={abs/2311.07362},
  url={https://api.semanticscholar.org/CorpusID:265150082}
}

@inproceedings{Li2023blip2BL,
  title={BLIP-2: Bootstrapping Language-Image Pre-training with Frozen Image Encoders and Large Language Models},
  author={Junnan Li and Dongxu Li and Silvio Savarese and Steven C. H. Hoi},
  booktitle={International Conference on Machine Learning},
  year={2023},
  url={https://api.semanticscholar.org/CorpusID:256390509}
}

@inproceedings{Sutton1999PolicyGM,
  title={Policy Gradient Methods for Reinforcement Learning with Function Approximation},
  author={Richard S. Sutton and David A. McAllester and Satinder Singh and Y. Mansour},
  booktitle={Neural Information Processing Systems},
  year={1999},
  url={https://api.semanticscholar.org/CorpusID:1211821}
}

@inproceedings{Liu2023MitigatingHI,
  title={Mitigating Hallucination in Large Multi-Modal Models via Robust Instruction Tuning},
  author={Fuxiao Liu and Kevin Lin and Linjie Li and Jianfeng Wang and Yaser Yacoob and Lijuan Wang},
  year={2023},
  url={https://api.semanticscholar.org/CorpusID:259251834}
}

@article{Liu2023VisualIT,
  title={Visual Instruction Tuning},
  author={Haotian Liu and Chunyuan Li and Qingyang Wu and Yong Jae Lee},
  journal={ArXiv},
  year={2023},
  volume={abs/2304.08485},
  url={https://api.semanticscholar.org/CorpusID:258179774}
}

@article{Zhu2023MiniGPT4EV,
  title={MiniGPT-4: Enhancing Vision-Language Understanding with Advanced Large Language Models},
  author={Deyao Zhu and Jun Chen and Xiaoqian Shen and Xiang Li and Mohamed Elhoseiny},
  journal={ArXiv},
  year={2023},
  volume={abs/2304.10592},
  url={https://api.semanticscholar.org/CorpusID:258291930}
}

@article{Ye2023mPLUGOwlME,
  title={mPLUG-Owl: Modularization Empowers Large Language Models with Multimodality},
  author={Qinghao Ye and Haiyang Xu and Guohai Xu and Jiabo Ye and Ming Yan and Yi Zhou and Junyan Wang and Anwen Hu and Pengcheng Shi and Yaya Shi and Chenliang Li and Yuanhong Xu and Hehong Chen and Junfeng Tian and Qiang Qi and Ji Zhang and Feiyan Huang},
  journal={ArXiv},
  year={2023},
  volume={abs/2304.14178},
  url={https://api.semanticscholar.org/CorpusID:258352455}
}

@inproceedings{Irvin2019CheXpertAL,
  title={CheXpert: A Large Chest Radiograph Dataset with Uncertainty Labels and Expert Comparison},
  author={Jeremy A. Irvin and Pranav Rajpurkar and Michael Ko and Yifan Yu and Silviana Ciurea-Ilcus and Chris Chute and Henrik Marklund and Behzad Haghgoo and Robyn L. Ball and Katie S. Shpanskaya and Jayne Seekins and David Andrew Mong and Safwan S. Halabi and Jesse K. Sandberg and Ricky Jones and David B. Larson and C. Langlotz and Bhavik N. Patel and Matthew P. Lungren and A. Ng},
  booktitle={AAAI Conference on Artificial Intelligence},
  year={2019},
  url={https://api.semanticscholar.org/CorpusID:58981871}
}

@article{Wang2022AutomatedRR,
  title={Automated Radiographic Report Generation Purely on Transformer: A Multicriteria Supervised Approach},
  author={Zhanyu Wang and Hongwei Han and Lei Wang and Xiu Li and Luping Zhou},
  journal={IEEE Transactions on Medical Imaging},
  year={2022},
  volume={41},
  pages={2803-2813},
  url={https://api.semanticscholar.org/CorpusID:248514098}
}

@inproceedings{li2022blip,
  title={Blip: Bootstrapping language-image pre-training for unified vision-language understanding and generation},
  author={Li, Junnan and Li, Dongxu and Xiong, Caiming and Hoi, Steven},
  booktitle={International Conference on Machine Learning},
  pages={12888--12900},
  year={2022},
  organization={PMLR}
}

@article{Liu2019ClinicallyAC,
  title={Clinically Accurate Chest X-Ray Report Generation},
  author={Guanxiong Liu and Tzu-Ming Harry Hsu and Matthew B. A. McDermott and Willie Boag and Wei-Hung Weng and Peter Szolovits and Marzyeh Ghassemi},
  journal={ArXiv},
  year={2019},
  volume={abs/1904.02633},
  url={https://api.semanticscholar.org/CorpusID:102353285}
}

@article{Li2020AuxiliarySK,
  title={Auxiliary signal-guided knowledge encoder-decoder for medical report generation},
  author={Mingjie Li and Fuyu Wang and Xiaojun Chang and Xiaodan Liang},
  journal={World Wide Web},
  year={2020},
  volume={26},
  pages={253 - 270},
  url={https://api.semanticscholar.org/CorpusID:219530927}
}

@article{Wang2018TieNetTE,
  title={TieNet: Text-Image Embedding Network for Common Thorax Disease Classification and Reporting in Chest X-Rays},
  author={Xiaosong Wang and Yifan Peng and Le Lu and Zhiyong Lu and Ronald M. Summers},
  journal={2018 IEEE/CVF Conference on Computer Vision and Pattern Recognition},
  year={2018},
  pages={9049-9058},
  url={https://api.semanticscholar.org/CorpusID:8285940}
}

@article{Chen2020GeneratingRR,
  title={Generating Radiology Reports via Memory-driven Transformer},
  author={Zhihong Chen and Yan Song and Tsung-Hui Chang and Xiang Wan},
  journal={ArXiv},
  year={2020},
  volume={abs/2010.16056},
  url={https://api.semanticscholar.org/CorpusID:226222210}
}

@article{touvron2023llama,
  title={Llama: Open and efficient foundation language models},
  author={Touvron, Hugo and Lavril, Thibaut and Izacard, Gautier and Martinet, Xavier and Lachaux, Marie-Anne and Lacroix, Timoth{\'e}e and Rozi{\`e}re, Baptiste and Goyal, Naman and Hambro, Eric and Azhar, Faisal and others},
  journal={arXiv preprint arXiv:2302.13971},
  year={2023}
}

\appendix

\label{sec:appendix}
\section{Appendix}
\subsection{More Cases.} 
\label{sec:More Cases}
\begin{figure*}[htbp]
\centering
\includegraphics[width=\textwidth]{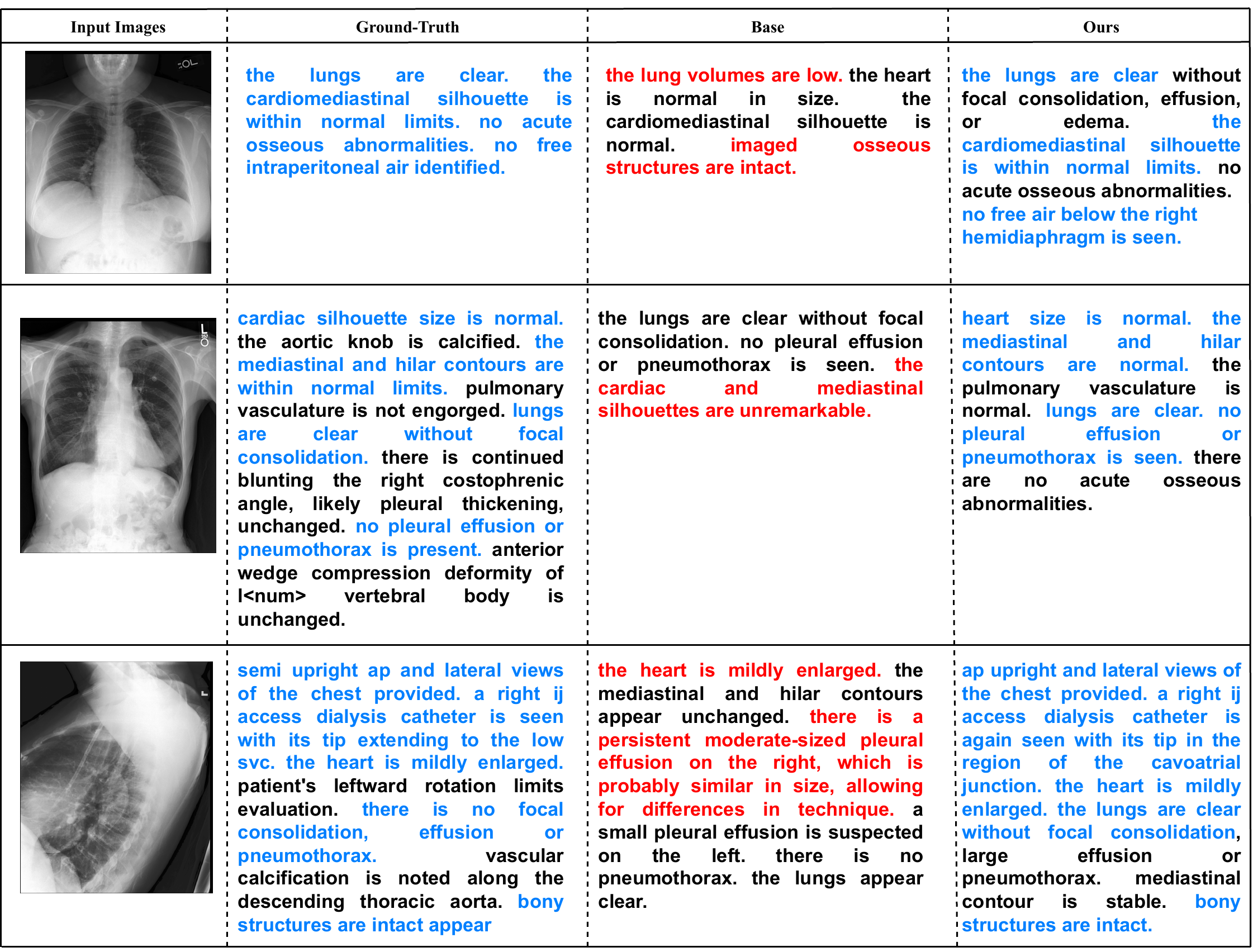}
\caption{ Qualitative examples of ground truth, ours and Base. Blue font indicates consistent content with the
ground-truth while red font indicates hallucinations. } \label{morecase}
\end{figure*}

More cases can seen in Figure~\ref{morecase}.
% \vspace{-0.8cm}

\end{document}